\long\def\comment#1{} 

\documentclass{article}
\usepackage{amssymb}
\usepackage{xcolor}
\usepackage{graphicx}

\title{{\bf A Tractable Inference Algorithm 
for Diagnosing Multiple Diseases\thanks{This work was supported by the NSF under 
Grant IRI-8703710, and by the NLM under Grant R01-LM04529.}}} 
 
\author{{\bf David Heckerman}\\ 
Medical Computer Science Group\\ 
Knowledge Systems Laboratory\\  
Departments of Computer Science and Medicine\\ 
Stanford, California 94305} 
 
\begin{document} 
 
\maketitle 
 
\begin{abstract} 
\noindent We examine a probabilistic model for the diagnosis of multiple
diseases.  In the model, diseases and findings are represented as
binary variables.  Also, diseases are marginally independent, features
are conditionally independent given disease instances, and diseases
interact to produce findings via a noisy {\sc or}-gate.  An algorithm
for computing the posterior probability of each disease, given a set
of observed findings, called {\em quickscore}, is presented.  The time
complexity of the algorithm is ${\cal O}(n m^{-} 2^{m^{+}})$, where
$n$ is the number of diseases, $m^{+}$ is the number of positive
findings and $m^{-}$ is the number of negative findings.  Although the
time complexity of quickscore is exponential in the number of positive
findings, the algorithm is useful in practice because the number of
observed positive findings is usually far less than the number of
diseases under consideration.  Performance results for quickscore
applied to a probabilistic version of Quick Medical Reference (QMR)
are provided.

\bigskip

\noindent 
Corrections to the original text are in \textcolor{red}{red}. Thanks to
Max Zhao for finding the error.

\end{abstract}

\section{Introduction} 
 
One of the most common criticisms of the use of probability theory in
expert systems is that the theory is impractical to apply in realistic
situations \cite{Buchanan84}.  In attempts to answer this criticism,
several researchers at Stanford, myself included, have undertaken the
task of converting to a probabilistic framework Quick Medical
Reference (QMR) \cite{Miller87}, one of the largest medical expert
systems in existence.
 
We have made a straightforward and tractable transformation of the QMR
knowledge base to a probabilistic
model \cite{Heckerman86a,Henrion88,Shwe90b}.  Like the heuristic
algorithms in QMR, the model allows for any combination of diseases to
be present in a patient.  Unfortunately, this feature leads to a time
complexity for inferring the probability of each disease given a set
of findings that is exponential in the number of {\em diseases}
($O(2^{n})$) for all known algorithms.  As there are over $600$
diseases in QMR, these algorithms are intractable.  Furthermore, this
problem is known to be NP-hard \cite{Cooper87a}.
 
In this paper, I present {\em quickscore}, an algorithm with time
complexity that is exponential in the number of {\em positive
findings} (findings observed to be present rather than absent).  More
precisely, the algorithm has a time complexity of $O(n m^{-}
2^{m^{+}})$, where $m^{+}$ is the number of positive findings and
$m^{-}$ is the number of negative findings.  Although quickscore has
an exponential time complexity, it is useful in practice because the
number of observed positive findings is often far less than the number
of diseases under consideration.  For many realistic patient cases,
quickscore implemented on a Macintosh II produces an answer in less
than 1 minute of real time.

\section{The QMR model} 
 
\begin{figure} 
\begin{center} 
\leavevmode 
\includegraphics[width=4.0in]{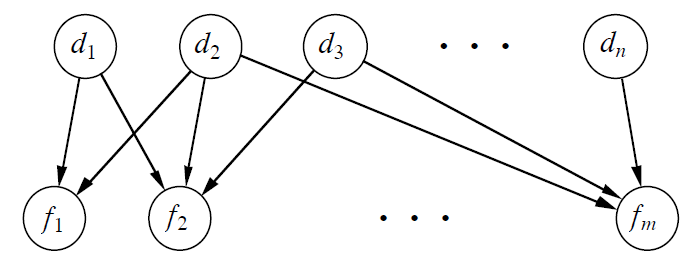}
\end{center}
\caption{A belief network for diagnosing multiple diseases. 
Diseases are marginally independent.  Findings are conditionally
independent given a disease instance.}
\label{fig:ci} 
\end{figure}

The probabilistic version of QMR is called QMR-DT for
Decision-Theoretic QMR. A belief network for QMR-DT, is shown in
Figure~\ref{fig:ci}.  Each of the $n$ nodes in the upper layer of the
network represents a disease that may be present or absent in a
patient.  Each of the $m$ nodes in the lower layer represents a
finding that may be observed to be present or absent in the patient,
or that may not be observed at all.  The problem of interest is to
compute the probability of each disease given a set of positive and
negative findings.
 
As indicated by the network in Figure~\ref{fig:ci}, we assume diseases
to be marginally independent.  Also, we assume that findings are
conditionally independent given any instance of the set of diseases.
An {\em instance of a set of diseases} is an assignment of present or
absent to each disease in that set.
 
\begin{figure} 
\begin{center} 
\leavevmode 
\includegraphics[width=2.0in]{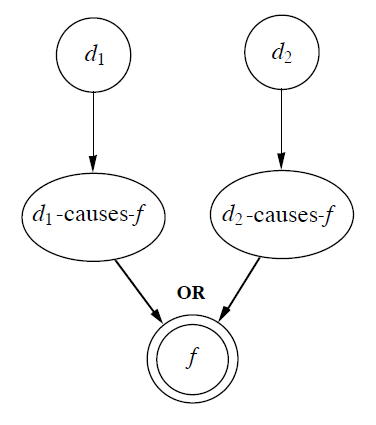}
\end{center}
\caption{A belief network for the noisy {\sc or}-gate. 
If disease $d_{1}$ is present, then with some probability it will act to cause 
finding $f$.  This probability depends neither on the state of $d_{2}$ nor on 
whether $d_{2}$ acts to cause $f$.  A reciprocal relationship holds for disease 
$d_{2}$.  If either disease acts to cause $f$, then $f$ will be observed to be 
present.}
\label{fig:noisy-or} 
\end{figure}

We also assume that diseases act independently to cause any given
finding to be present.  A belief network that represents this
independency for two diseases is shown in Figure~\ref{fig:noisy-or}.
The node labeled $d_{i}$--causes--$f$ represents the event that
$d_{i}$ causes finding $f$ to be present. The node with the double
boundary, labeled $f$, is a deterministic node that says finding $f$
will be present if and only if $d_{1}$ causes $f$ to be present,
$d_{2}$ causes $f$ to be present, or both $d_{1}$ and $d_{2}$ cause
$f$ to be present. The arc between the node labeled $d_{1}$ and the
node labeled $d_{1}$--causes--$f$ reflects our assumption that the
presence or absence of $d_{1}$ influences the probability that $d_{1}$
causes $f$ to be present.  In particular, we assume that, if $d_{1}$
is present, it may cause $f$ to be present with some probability.  If
$d_{1}$ is absent, we assume that the disease cannot act to cause $f$.
The same set of assumptions holds for the disease $d_{2}$.  Finally,
the lack of arcs between nodes in the upper two rows of the figure
reflects the assertions that (1) the probability distribution for the
variable $d_1$--causes--$f$ depends neither on the absence or presence
of $d_2$ nor on the absence or presence of the event
$d_2$--causes--$f$, and (2) the probability distribution for the
variable $d_2$--causes--$f$ depends neither on the absence or presence
of $d_1$ nor on the absence or presence of the event
$d_1$--causes--$f$.  We call this form of conditional independence
{\em causal independence} to distinguish it from the type of
conditional independence that is commonly represented in belief
networks.
 
Under these assumptions, we can compute the probability of $f$ given
any instance of the pair $\{d_{1},d_{2}\}$ from the two assessments
$p_{1}$ and $p_{2}$, where $p_{i}$ is the probability that $d_{i}$
causes $f$, $i = 1,2$.  If both diseases are absent, finding $f$ must
be absent.  If only one disease is present---say, $d_{i}$---then
\begin{displaymath}  
p(f^{+}|{\rm only} \ d_{i}) = p_{i} 
\end{displaymath} 
and 
\begin{displaymath} 
p(f^{-}|{\rm only} \ d_{i}) = 1 - p_{i} 
\end{displaymath} 
where $f^{+}$ and $f^{-}$ denote the presence and absence of finding $f$, 
respectively.  That is, the probability that $d_{i}$ causes $f$ is just the 
probability of $f$ given that only disease $d_{i}$ is present.  If both $d_{1}$  
and $d_{2}$ are present, finding $f$ will be absent only if both diseases fail 
to 
cause $f$ to be present.  Therefore,   
\begin{displaymath} 
p(f^{-}|d_{1}^{+},d_{2}^{+}) = (1 - p_{1}) (1 - p_{2}) = p(f^{-}| {\rm only} \  
d_{1}) p(f^{-}|{\rm only} \ d_{2})  
\end{displaymath} 
where $d_{i}^{+}$ denotes the presence of disease $d_{i}$. 
 
More generally, suppose that $n$ diseases can potentially cause
finding $f$, and that these $n$ diseases are the only causes of $f$.
As in the simpler case,
\begin{displaymath}  
p(f^{+}|{\rm only} \  d_{i}) = p_{i} 
\end{displaymath} 
where $p_{i}$ is the probability that $d_{i}$ causes $f$.  Let $D_{i}$ denote a 
particular instance of the $n$ diseases and let $D_{i}^{+}$ denote the set of 
diseases that are present in the instance $D_{i}$. Given instance $D_{i}$, 
$f$ will be absent only if none of the diseases in $D^{+}_{i}$ cause 
$f$ to be present.  It follows that 
\begin{equation} 
p({f}^{-}|{D}_{i}) = \prod_{d \in D_{i}^{+}} p({f}^{-}|{\rm only} \  d) 
\label{eq:noisy-or} 
\end{equation} 
Thus, using Equation~\ref{eq:noisy-or}, we can compute the probability
of $f$ given any of the $2^{n}$ disease instances from only $n$
assessments.
 
The assumptions underlying Equation~\ref{eq:noisy-or}, including the
assumption of causal independence, have been described
previously \cite{Good61}, and several researchers have suggested that
these assumptions be used to model medical domains
\cite{Habbema76,Peng86b}.  Pearl has called this canonical model of cause 
and effect the {\em noisy {\sc or}-gate} \cite{Kim83,Pearl86b}.  Other
researchers have extended the model to accommodate situations where a
finding may appear in the absence of all explicitly represented
diseases, a leaky {\sc or}-gate \cite{Suppes70,Henrion87b}.
 
The leaky {\sc or}-gate model was assumed implicitly by the developers
of the QMR expert system \cite{Miller87}.  In conjunction with the
independence assumptions shown in Figure~\ref{fig:ci}, the model made
tractable the transformation of QMR into a probabilistic framework.
As we see in the following section, these assumptions also accommodate
an inference algorithm that is tractable for many patient cases.  In
Section~\ref{sec:relax}, we examine the appropriateness of these
assumptions, and discuss how they might be relaxed.

\section{The Quickscore Algorithm} 
 
The goal of quickscore is to compute the probability of each disease
$d_{i}$, $i~=~1,2,~\ldots~n$, given a set of positive findings $F^{+}$
and a set of negative findings $F^{-}$, under the assumptions
described in the previous section.\footnote{Here, we develop the
algorithm for nonleaky {\sc or}-gates.  We can, however, extend easily
the development to include leaky {\sc or}-gates.}  To understand how
quickscore works, we first compute the probability that a single
finding $f$ will be absent.  Using the expansion rule, we
get \begin{equation} p(f^{-}) = \sum_{D_{k} \in D} p(f^{-}|D_{k})
p(D_{k}) \label{eq:1}
\end{equation} 
where $D$ is the set of all disease instances.  Using the assumption that 
diseases are marginally independent and the assumptions underlying the 
noisy {\sc or}-gate, Equation~\ref{eq:1} becomes 
\begin{equation} 
p(f^{-}) = \sum_{D_{k} \in D} \left[ \prod_{d \in D_{k}^{+}} p(f^{-}|{\rm only} 
\  d) 
\prod_{d \in D_{k}^{+}} p(d^{+}) \prod_{d \in D_{k}^{-}} p(d^{-}) \right]  
\label{eq:2}   
\end{equation} 
where $D_{k}^{-}$ denotes the set of diseases that are absent in the instance 
$D_{k}$.  \comment{(To simplify the discussion, we shall ignore leak terms in 
the  
noisy {\sc or}-gate.)} 
 
Now consider the expression  
\begin{equation} 
\prod_{i = 1}^{n} \left[ p(f^{-}|{\rm only} \  d_{i}) p(d_{i}^{+}) + p(d_{i}^{-
}) 
\right]  \label{eq:3} 
\end{equation} 
If we multiply out this expression, we see that it is just the right-hand side 
of 
Equation~\ref{eq:2}.   
Thus, we obtain 
\begin{equation} 
p(f^{-}) = \prod_{i = 1}^{n} \left[ p(f^{-}|{\rm only} \  d_{i}) p(d_{i}^{+}) + 
p(d_{i}^{-}) \right]  \label{eq:4} 
\end{equation} 
The difference in time complexity between the computations of Equations 
\ref{eq:2} 
and \ref{eq:4} is striking.  The computation in Equation \ref{eq:2} is
a sum over $2^n$ terms, whereas the computation in Equation \ref{eq:4}
is a linear product over $n$ sums.  Pearl \cite[pp. 187-188]{Pearl88}
was the first to note the equivalence between these two computations.
As we shall see, quickscore derives its speed from this equivalence.
 
Under the assumption that findings are conditionally independent given
any disease instance, we can employ the transformation described in
the previous paragraph to compute the probability that the set of
negative findings, $F^-$, are observed.  We have
\begin{equation} 
p(F^{-}) = \prod_{i = 1}^{n} \left( \left[ \prod_{f \in F^{-}} p(f^{-}| 
{\rm only} \  d_{i}) \right] p(d_{i}^{+}) + p(d_{i}^{-}) \right)  \label{eq:5} 
\end{equation} 
 
The situation is more complex for positive findings.  Let us first 
examine the simple case where $F^{+} = \{f_{1},f_{2}\}$.  Applying the expansion 
rule to  $p(f_{1}^{+},f_{2}^{+})$ and using the assumption of conditional 
independence of findings, we get  
\begin{equation} 
p(f_{1}^{+},f_{2}^{+}) = \sum_{D_{k} \in D} p(f_{1}^{+}|D_{k}) 
p(f_{2}^{+}|D_{k})  
p(D_{k})  
\label{eq:1+} 
\end{equation} 
Since $p(f_{j}^{+}|D_{k}) = 1 - p(f_{j}^{-}|D_{k})$, Equation~\ref{eq:1+} 
becomes 
\begin{eqnarray} 
p(f_{1}^{+},f_{2}^{+}) & = & \sum_{D_{k} \in D} p(D_{k}) - \nonumber \\ 
& & \sum_{D_{k} \in D} p(f_{1}^{-}|D_{k}) p(D_{k}) - \nonumber \\ 
& & \sum_{D_{k} \in D} p(f_{2}^{-}|D_{k}) p(D_{k}) + \nonumber \\ 
& & \sum_{D_{k} \in D} p(f_{1}^{-}|D_{k})  p(f_{2}^{-}|D_{k}) p(D_{k}) 
\label{eq:2+}  
\end{eqnarray} 
The first sum in Equation~\ref{eq:2+} is equal to $1$.  The remaining
terms are in the same form as the right-hand side of
Equation~\ref{eq:1}.  Thus, using the algebraic transformations
derived previously, we obtain
\begin{eqnarray} 
p(f_{1}^{+},f_{2}^{+}) & = & 1 - \nonumber \\ 
& & \prod_{i = 1}^{n} \left[ p(f_{1}^{-}|{\rm only} \  d_{i}) p(d_{i}^{+}) + 
p(d_{i}^{-}) \right] - \nonumber \\ 
& & \prod_{i = 1}^{n} \left[ p(f_{2}^{-}|{\rm only} \  d_{i}) p(d_{i}^{+}) + 
p(d_{i}^{-}) \right] + \nonumber \\ 
& & \prod_{i = 1}^{n} \left[ p(f_{1}^{-}|{\rm only} \  d_{i})  
p(f_{2}^{-}|{\rm only} \  d_{i}) p(d_{i}^{+}) + p(d_{i}^{-}) \right] 
\label{eq:3+} 
\end{eqnarray} 
More generally, 
\begin{equation} 
p(F^{+}) = \sum_{F^{'} \in 2^{F^{+}}} (-1)^{|F^{'}|} \prod_{i = 1}^{n} \left( 
\left[  \prod_{f \in F^{'}} p(f^{-}|{\rm only} \  d_{i}) \right] p(d_{i}^{+}) + 
p(d_{i}^{-}) \right) \label{eq:4+} 
\end{equation} 
where $2^{F^{+}}$ denotes the power set of $F^{+}$, and $|F^{'}|$ denotes the 
number of elements in set $F^{'}$. 
 
In the most general case where some findings are present and some are
absent, we can combine Equation~\ref{eq:5} and Equation~\ref{eq:3+} to
obtain
\begin{equation} 
p(F^{+},F^{-}) = \sum_{F^{'} \in 2^{F^{+}}} (-1)^{|F^{'}|} \prod_{i = 1}^{n}  
\left( \left[  \prod_{f \in F^{'} \cup F^{-}} p(f^{-}|{\rm only} \  d_{i}) 
\right] 
p(d_{i}^{+}) + p(d_{i}^{-}) \right) \label{eq:1gen} 
\end{equation} 
It is now a simple matter to compute $p(d_{i}^{+}|F^{+},F^{-})$.  
First, we compute 
$p(F^{+},F^{-})$.  Then, we compute \textcolor{red}{$p(F^{+},F^{-}|d_{i}^{+})$} by 
setting $p(d_{i}^{+}) = 1$ and $p(d_{i}^{-}) = 0$ in Equation~\ref{eq:1gen}.  
The sought-after probability is then 
\begin{equation} 
p(d_{i}^{+}|F^{+},F^{-}) = 
\frac{\textcolor{red}{p(F^{+},F^{-}|d_{i}^{+}) \ p(d_{i}^{+})}}{p(F^{+},F^{-})}.
\label{eq:Bayes} 
\end{equation} 
\comment{Note that we do not have
to compute from scratch both terms on the right-hand side of
Equation~\ref{eq:Bayes}.  When computing $p(F^{+},F^{-})$, we can
cache intermediate results, and then use these results to obtain
$p(d_{i},F^{+},F^{-})$.}
 
The quickscore algorithm can provide intermediate results.  In
particular, suppose that we order the findings in $F^{+}$---say,
$f_{1}$, $f_{2}$, \ldots $f_{m^{+}}$.  In Equation~\ref{eq:1gen}, we
can first compute the term in the power set of $F^{+}$ corresponding
to $f_{1}$ alone.  We can then compute the terms in the power set that
correspond to combinations of only $f_{1}$ and $f_{2}$.  Continuing in
this way, the probability of each $d_{i}^{+}$ given the first $j$ findings
can be recovered from the algorithm at any time, where time is an
exponential function of $j$.  The fact that quickscore can provide
intermediate results may prove useful.  The QMR knowledge base
contains a partial ordering of findings by their general clinical
importance in determining a diagnosis.  This ordering might be used to
degrade gracefully the performance of quickscore with increasing time
constraints.

\section{Run-Time Performance of Quickscore} 
 
\begin{figure} 
\begin{center} 
\leavevmode 
\includegraphics[width=4.0in]{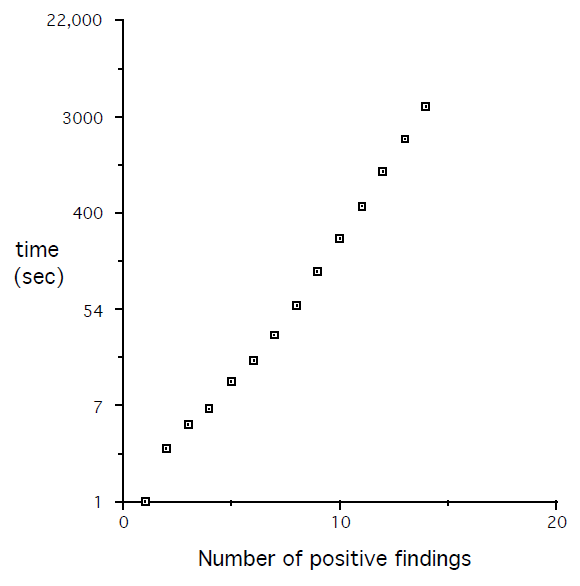}
\end{center}
\caption{Run times for quickscore.
Run times for quickscore implemented in LightSpeed Pascal on a Macintosh II are 
plotted against the number of positive findings in a case.  Nine positive 
findings  
typically can be scored in less than 1 minute.}
\label{fig:run-times} 
\end{figure}

Quickscore has been implemented in Lightspeed Pascal on the Macintosh
II computer.  Figure~\ref{fig:run-times} shows the run time of the
algorithm for cases of various size.  The cases are taken from a
library of classic cases used by the QMR research team to test
periodically the diagnostic accuracy of the heuristic knowledge base.
These cases contain only positive findings.  As the graph indicates,
nine findings can typically be scored in less than 1 minute.  Overall,
in 25 percent of the 400 cases in the library, quickscore requires 15
minutes or less to score each case.
 
\section{Weaknesses of the Algorithm} 
\label{sec:relax} 
 
The diagnostic model described in this paper contains several
assumptions that may not be appropriate for many medical and
nonmedical domains.  For example, some diseases and findings can occur
with different degrees of severity, and hence are not two-valued.  In
addition, certain diseases cause others to be present.  Consequently,
all diseases are not marginally independent.  Furthermore, some
findings are conditionally dependent, and certain diseases are caused
by findings rather than vice-versa (e.g., a history of alcoholism
tends to cause cirrhosis of the liver).  None of these observations
pose a significant barrier to the translation of the QMR's knowledge
base to a probabilistic framework.  For example, several researchers
have generalized the noisy {\sc or}-gate model, constructing other
prototypic models that embody the assumption of causal
independence \cite{Kim83,Heckerman88,Henrion87b}.  We can use these
prototypic models to accommodate multiple-valued diseases and
findings, and to accommodate causal interactions among diseases.
Also, by introducing hidden or unobservable pathophysiologic states,
we can capture many of the dependencies among findings.  Again, we can
use the noisy {\sc or}-gate and its extensions to model interactions
between diseases and pathophysiologic states.  Unfortunately, we
cannot extend quickscore in a straightforward manner to treat these
extensions to the current QMR-DT model.
 
In addition, quickscore is unstable numerically when
$p(F^+,F^-)$---the probability of observations for a given case---is
small, because $p(F^+,F^-)$ is a sum of positive and negative terms on
the order of $10^0$ (see Equation~\ref{eq:1gen}).  IEEE-standard
extended-precision arithmetic on the Macintosh II provides about 19
decimal places of precision.  Thus, inferences using the current
implementation of quickscore (or any implementation on hardware that
uses the IEEE standard) are unreliable when $p(F^+,F^-)$ is less than
approximately $10^{-19}$.  Typically, for the current QMR-DT model,
such situations arise when the number of positive findings exceeds 15.

\section{Conclusion} 
 
Despite its shortcomings, quickscore has been useful in the
development of QMR- DT.  For example, the QMR-DT group recently has
experimented with several approximation algorithms for inference that
are based on Monte-Carlo techniques \cite{Shwe90b}. These algorithms
are suited to inference in the extensions of the QMR-DT model
discussed in the previous section, but their convergence properties
are not well characterized.  Quickscore has provided gold standards
for evaluating such convergence properties in the context of the
current model.  In general, quickscore is likely to be a useful tool
for knowledge engineers who develop decision-theoretic expert systems.

\section{Acknowledgments} 
 
I thank Gregory Copper for encouraging me to implement the quickscore
algorithm, Michael Shwe for assisting with the implementation, the
entire QMR-DT group at Stanford for useful discussions, and Lyn Dupre
for reviewing several drafts of this manuscript.
  
\bibliographystyle{apalike} 

\bibliography{david} 
 
\end{document}